# Context Specific Event Model For News Articles

[1] Kowcika A, [2] Uma Maheswari, [3] Geetha T V

[1] Department of Computer Science and Engineering, College of Engineering Guindy
Anna University, Chennai.TamilNadu-600025, India

[2] Department of Computer Science and Engineering, College of Engineering Guindy
Anna University, Chennai.TamilNadu-600025, India

[3] Department of Computer Science and Engineering, College of Engineering Guindy
Anna University, Chennai.TamilNadu-600025, India

**Abstract**
We present a new context based event indexing and event ranking model for News Articles. The context event clusters formed from the UNL Graphs uses the modified scoring scheme for segmenting events which is followed by clustering of events. From the context clusters obtained three models are developed- Identification of Main and Sub events; Event Indexing and Event Ranking. Based on the properties considered from the UNL Graphs for the modified scoring main events and sub events associated with main-events are identified. The temporal details obtained from the context cluster are stored using hashmap data structure. The temporal details are place-where the event took; person-who involved in that event; time-when the event took place. Based on the information collected from the context clusters three indices are generated- Time index, Person index, and Place index. This index gives complete details about every event obtained from context clusters. A new scoring scheme is introduced for ranking the events. The scoring scheme for event ranking gives weight-age based on the priority level of the events. The priority level includes the occurrence of the event in the title of the document, event frequency, and inverse document frequency of the events.

## 1. Introduction

Event extraction is a particularly challenging type of information extraction (IE). Information retrieval systems are responsible to provide the information of interest to users. It is the process of extracting the structured information from unstructured text. The system in [1] states that IE systems were evaluated by the Message Understanding Conferences (MUC) till 1998. Automatic Content Extraction (ACE) program is the successor of MUC with the objective of developing the extraction technology to support automatic processing of source language data. This includes classification, filtering, and selection based on the language content of source data, i.e., based on meaning conveyed by the data.

In [7] ACE defines three basic kinds of information to be extracted from the natural language text such as entities, relations and events.And also the system [7] talks about the number of properties of events namely Polarity, Tense, Genericity and Modality which are related to when, where and if the event really took place etc. Once the event of interest is identified, event information is added as metadata to the text document. ACE defines a process of identifying events from only single sentence.

Event extraction is one of the challenging research points in information extraction. The goal of event extraction is to describe an event using natural language to predict the time, place and other participants and actions about an event. Event extraction can be used in many NLP application fields, such as automatic summarization discussed in [3], question and answering discussed [2], and information retrieval discussed [2] and so on.

The system in [8] says that temporal information extraction is a subtask of information extraction (IE). Its goal is to extract time expressions and temporal relations from natural language text and its representation. Processing temporal information in natural language has its value in natural language processing (NLP) tasks. For example, temporal information processing is crucial in the temporal question and answering systems. To answer a "when" question the system needs to temporally anchor the event, and to answer a "how long" question the system needs to measure the duration of the event and "why" the system needs the reason behind the event.

The internet contains more than thousands of electronic collections that often contain high quality information. The system [4] talks about the basic aim of selecting the best collection of information for particular information need. The indexing phase of search engines can be viewed as a Mining of web content. Starting from a collection of unstructured documents, the indexer extracts a large amount of information like the list of documents, which contain a given term and other details like number of all the occurrences of each term within every document. This information is maintained in an index, which is usually represented using an inverted file (IF) which is the most widely adopted format for this index due to its efficiency of its usage. The index consists of an array of the posting lists and contains the term as well as the identifiers of the documents containing the term. The term based is less efficient. Thus the significance of term for building the index is reduced and the research laid on the context of the document. Context provides extra information to improve search result relevance. The context of a document cab be easily derived using the relations extracted from UNL Graphs.

An event in two news stories can be defined as a specific happening at a certain time, in a specific place and involves two or more number of participants. Different news articles talks about the same event in different perspective. It is interesting and challenging to gather information about same or similar events from news corpus.

The study in [20] talks about ranking events from documents is mainly present in automatic summarization. If different events contain the same element, these different events have associative relations between these events. Previous approaches had been used this kind of event relations to construct event map for a document and compute event importance using Page Rank algorithm. There are two problems about the existing methods. First, it is very hard to extract elements for every event elements. Second, the associative strength of events is different and it is not accurate to depict event relation.

## 2. Related Work

N.McCracken et al [5] combined statistical and knowledge based technique for extracting events. It mainly focuses on the summary report genre. He focuses on developing a system that allows the utilization of statistical techniques without new training data.

F.Xu et al [9] developed a methodology for identifying event extent, event trigger and event argument automatically. This work extracted the events from the Nobel Prize winning domain by obtaining extraction rules using binary relations. This method extracts the events found in every sentence. It does not look for the events that have its scope in more than one sentence. Salem Abuleil [6] proposed a method that can extract events by breaking each event into elements analyzes and understands the syntax of each element, identifies the role played by each element in the event and how they form relationship between events.

C.Aone et al [10] identifies events by tagging the text and used pattern matching techniques and rule based approach. It does not perform the complete analysis of semantics. All the above said existing systems extract the events without considering the meaning of the text and it looks only for content and not for context. However, consideration of meaning and context of the text improves the efficiency of event extraction, and the information extraction as a whole.

Riloff [13] initiated and claimed that if a corpus can be divided into documents involving a certain event type and those not involving that type, patterns can be evaluated based on their frequency in relevant and irrelevant documents. Yangarber et al. [14] incorporated Riloff's metric into a bootstrapping procedure, which started with several patterns but required no manual document classification or annotation. The patterns were used to identify some relevant documents, and the top-ranked patterns were added to groups. This process was repeated, assigning a relevance score to each document based on relevance of the patterns it contains and gradually growing the set of relevant patterns.

In [15], the authors introduce a double indexing mechanism for search engines based on campus Net which is based on full-text search engine, but it is a private net. The CNSE has crawl machine, Chinese automatic segmentation, and index and search machine. They proposed double apple indexing mechanism, which has both document index and word index. In the retrieval, the search engine first gets the document id of the word in the word index, and then goes to the position of that particular word in document index. Because in the document index, the word in the same document is adjacent, the search engine directly compares the largest word matching assembly with sentence that users give. The mechanism proposed by them seems to be time consuming as the index exists at different levels.

Another work described was the reordering algorithm in [16] which partitions the set of documents into some ordered clusters on the basis of similarity measure. According to this algorithm, the biggest document is selected as centroid of the first cluster and most similar documents are assigned to the cluster. Then the biggest document is selected and same process repeats. This algorithm is not effective in clustering the most similar documents together. The biggest document may not have similarity with any of other documents but still it is taken as the representative of the cluster.

Another proposed work was the threshold based clustering algorithm [19] in which the number of clusters is not known. However, two documents are classified to the same cluster if the similarity between them according to the specified threshold. This threshold is defined by the user before starting the algorithm. It is easy to see that if the threshold is small then all the elements will get assigned to different clusters. If the threshold is large then the elements may get assigned to just one cluster. Thus the algorithm is sensitive to specification of the threshold. Stevenson and Greenwood [11] proposed an alternative method for ranking the candidate patterns. They had used WordNet to calculate word similarity and had chosen vector to represent each pattern. Later, Greenwood and Stevenson [12] introduced a structural similarity measure that could be applied to the extraction patterns consisting of the linked dependency chains.

Zhong and Liu [20] take events as the basic semantic unit for texts to study the method of identifying events and ranking event for a single document. The key technique is based on the analysis of event relations to construct the event relation graph as the representation model for a single document, further applying PageRank algorithm to compute the event weight.

The paper [18] deals two kinds of bootstrapping methods used for event extraction they are the document-centric and similarity-centric approaches, and proposes a filtered ranking method that combines the advantages of the two methods. They analyze the results using two evaluation metrics and observe the effect of different training corpus. Their experiments show that his ranking method achieves higher performance on different evaluation metrics and stable across different corpus.
.
## 3. Architecture

In this work, the input to the clustering algorithm is information segments obtained from the document after semantic representation. The underlying semantic representation used is the language independent UNL (Universal Networking Language) representation [21]. In UNL representation sentences are represented by an UNL graph consisting of UNL concepts with edges indicating relations between concepts. However in this work sub-segments (concept-relation-concept) of the UNL graph [21] corresponding to sentence constituents are considered. Therefore the input to the clustering algorithm are UNL sub graphs which can be Concept-Relation-Concept(C-R-C), Concept(C)-Relation(R) and Concept(C) only. In effect we are dealing with graph based clustering of UNL based semantic sub graphs representing sentence constituents.

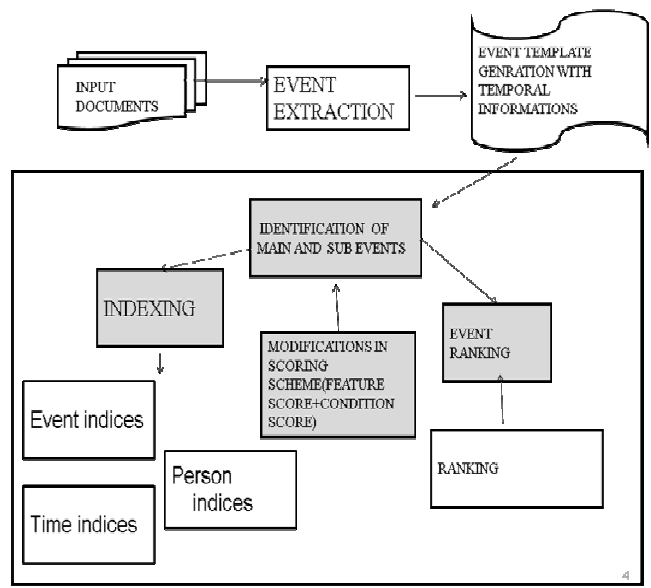

Fig.1 Architecture

The system [22] used a new scoring scheme for identifying event specific sentences. Each sentence is checked with conditions and scores are added according to the similarity of the sentences. The probability values between the sentences are obtained. The sentences with maximum probability value are grouped under that particular event. Multiple events will be obtained for each document. Event specific clustering is performed. The same scoring scheme is used for clustering event specific sentences. This is the inter-document clustering where events from multiple documents are clustered using the scoring scheme.

The segments with same probability value are clustered under that particular event. Multiple events will be obtained from multiple documents. The proposed system adds some features to the existing scoring scheme of [22] for segmentation. The modified scoring scheme includes conjunction score along with the condition score and feature score for segmentation. The new improved scoring scheme improves the segmentation quality by grouping the continuous events under same segment. The highly improved segments are given as inputs for the clustering algorithm. The events clusters formed by default will be well-formed clusters are shown in Fig.2
.
The below algorithm describes the steps for performing a event clustering

3.1 Algorithm

**Input:** Improved Event Segments.
**Output:** Event Clusters.
**Algorithm:**
for(s= 1 to n) do  ---------------- Segments
 for(c=1 to m) do  ------------------Concepts
if(ci contains "icl>event")
p = 0.5;
else if(ci contains "icl>action")
p = 0.4;
 &&
if(cj contains "icl>place")
p1 =  0.2;
 &&
 if(ck contains "icl>person")
 p2 =  0.2;
 &&
sss
 if(cl contains "pos" as "dur")
 p3 = 0.1;
 Similarity Score = Condition Score + Feature Score (All Feature)
 S = p + p1 + p2 + p3;
 if( S > 0.8)
Form Event Clusters.

{Condition Score}
{Feature Score}

Fig.2. Context Based Event Clusters

## 4. Identification of Main Sub-Events

From the clusters obtained the temporal details are stored using hashmap data structure. The temporal details are place-where the event took; person-who involved in that event; time-when the event took place. The template will fill the empty slots for temporal details with the user-specified query. The template will displays the sub-events associated with the main events. The sub-events also apply the same improved scoring scheme according to its properties considered.

The details stored in the hashmap are head nodes, concept nodes, relations between the concept nodes, frequency, pos tagging, document id. This information is extracted from clusters obtained from clustering.Fig.3.displays the identification of sub-events related to the main events.

4.1 Properties used for identification of Main-events

* Temporal Expressions ( Place, Time, Location)
* UNL constraints
* POS tagging from UNL Graphs
* Frequency of the concepts
* Rules for timeline calculations

4.2 Properties used for identification of Sub-events

* Temporal Expressions (Place, Time, Location)
* UNL constraints

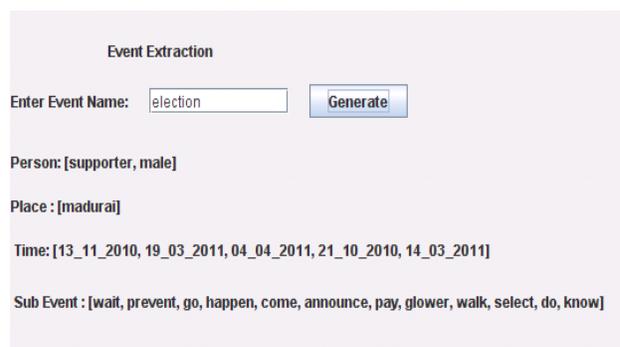

Fig.3.Identification of Main and Sub Events

## 5. Event Indexing

The purpose of storing an index is to optimize speed and performance in finding relevant documents for the search query. Without an index, the search engine would scan every document in the corpus, which would require considerable time and computing power. Indexing collects, parses, and stores data to facilitate fast and accurate information retrieval.

The proposed system introduces three indices namely Event index, Person Index, Place index. Person Indices consists of the following fields Person, Event name, Sub-event, Document ID, Place, Time, and Sentence ID. In Person Index we find for the person details that in what are all the events he involved in and what are the sub-events connected with it and location of the event took and the document ID where the events description is given and also the sentence ID.Fig.4 displays the person index developed using the context specific approach of the above mentioned improved scoring scheme.

Place Indices consists of the following fields Place, Event name, Sub-event, Document ID, Person, Time, and Sentence ID. In Place Index we find for the place details that in what are all the events happened in that place and what are the sub-events connected with it and location of the event took and the document ID where the events description is given and also the sentence ID.Fig.5 displays the place index developed using the context specific approach of the above mentioned improved scoring scheme.

Event Indices consists of the following fields Event name, Sub-event, Document ID, Person, Place, Time, and Sentence ID. In Event Index we find for the event details that events happened in which place and what are the sub-events connected with it and location of the event took and the persons involved in that event the document ID where the events description is given and also the sentence ID.Fig.6 displays the event index developed using the context specific approach of the above mentioned improved scoring scheme.

### 5.1 List of properties for Indexing

- Main and sub event tagging(based on scores)
- Time Tagging(for vague expressions-it can be tackled based on the UNL attributes, constraints)
- Frequency of Persons (Number of persons involved in each event)
- Number of places the event occurred

| Place | Event_Name | Sub_Event | Document_ID | Person | Time | Sentence |
|---|---|---|---|---|---|---|
| Madurai | election | [Do,know] | Ta_malar3 | [supporter,male] | 14_03_2011 | [s1,s2] |
| Madurai | election | [wait,prevent,go,happen,co..] | Ta_bbc4 | [supporter,male] | 13_11_2010 | [s1,s2,s3,s4,s5,s6,s7...] |
| Madurai | incident | [fulfill] | Ta_thanthi5 | [singer,devotee,minister,poli...] | 21_10_2010 | [s1,s2,s3,s4,s5,s6,s7...] |
| Madurai | incident | [dance] | Ta_thanthi8 | [singer,devotee,minister,poli...] | 01_05_2010 | [s1,s2,s3,s4,s5,s6,s7...] |
| Madurai | incident | [do,talk,wait,know] | Ta_malai9 | [singer,devotee,minister,poli...] | 26_10_2010 | [s1,s2,s3,s4,s5,s6,s7...] |
| Madurai | incident | [] | Ta_thinam6 | [singer,devotee,minister,poli...] | 04_04_2011 | [s2,s3,s4,s5,s6,] |
| Madurai | incident | [happen.,wait] | Ta_malai1 | [singer,devotee,minister,poli...] | 17_07_2010 | [s1,s2,s3,s4,s5,s6] |

Fig. 5 Place Index

| Event_Name | Sub_event | Document_id | Persons | Places | Time | senetence |
|---|---|---|---|---|---|---|
| election | [] | Ta_thanthi1 | [Supporter,male] | [Madurai] | 04_04_2011 | [s2,s3,s4,s5,s6] |
| election | [pay,go,wait,glower,walk] | Ta_malar1 | [Supporter,male] | [Madurai] | 19_03_2011 | [s1,s2,s3,s4,s5,s6] |
| election | [Select] | Ta_malai4 | [Supporter,male] | [Madurai] | 21_10_2010 | [s1,s2] |
| election | [do,know] | Ta_mama10 | [Supporter,male] | [Madurai] | 14_03_2011 | [S1,s2] |
| election | [wait,prevent,go,happen,co....] | Ta_bbc4 | [Supporter,male] | [Madurai] | 13_11_2010 | [s1,s2,s3,s4,s5,s6,s7...] |

Fig.6.Event Index

## 6. Event Ranking

Several researchers have proposed semi-supervised learning methods for adapting event extraction systems to new event type models. The proposed system introduced a new approach for ranking. It uses the scoring method for ranking. Scoring is based on the priorities given for the following number of documents in which the events occur, event frequency for the document, and finally the weight age given to the occurrence of the event in the title. Then the scores are analyzed and for the event which score is higher their priority level is also higher and ranking is given in that order. Fig.6 displays the ranks that are calculated using the context specific approach of above mentioned improved scoring scheme.

6.1 Algorithm for Ranking:

**Input:** Event Clusters.

**Output:** Ranks.

| Person | Event_Name | Sub_Event | Document_ID | Places | Time | Sentence |
|---|---|---|---|---|---|---|
| Student | Competition | [Participate] | Ta_bbbc | [there, below] | 21_10_2010 | [s1,s2] |
| Student | Competition | [Happen, wait] | Ta_malar1 | [there,below] | 17_07_2010 | [s1,s2,s3,s4,s5,s6] |
| Student | Competition | [wait,do,jump,pay] | Ta_Malar2 | [there,below] | 21_10_2010 | [s1,s2,s3,s4,s5,s6...] |
| Student | Competition | [Devolpe,collide] | Ta_malar4 | [there,below] | 03_07_2010 | [s1,s2] |
| Student | Festival | [walk,begin] | Ta_malar5 | [area,mound opening] | 21_10_2010 | [s1,s2,s3,s4,s5,s6] |
| Student | Festival | [open,conduct,dipute,do] | Ta_bbc3 | [area,mound,opening] | 21_10_2010 | [s1,s2,s3,s4,s5,s6,s7..] |
| Student | Festival | [do,walk,happen] | Ta_Malar4 | [area,mound,opening] | 02_06_2010 | [s1,s2,s3,s4,s5,s6,s7] |

Fig.4 Person Index

```
if(the event is present in more no of documents)
{
Scores are added
}
if(event frequency in the document is higher)
{
Scores are added
}
if(document heading contains the event)
{
Scores are added
}
Finally all scores are added

Rank(Based on the scores)
```

## 7. Performance Evaluation

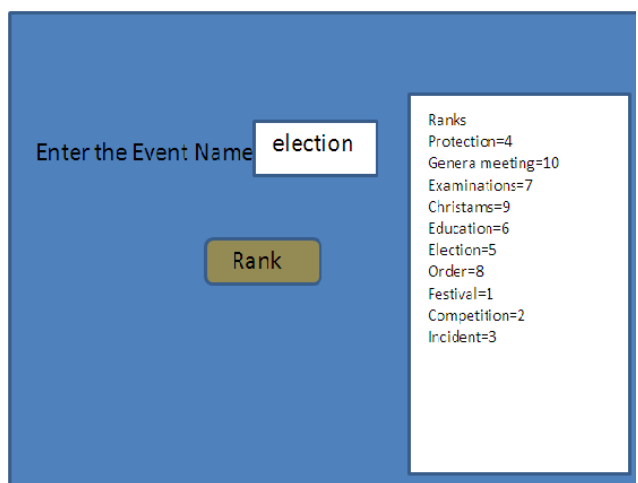

Fig 7. Content Based Event Indexing

### 7. 1 .Evaluation Parameter

The Evaluation parameter used for the proposed system is Silhouette Coefficient [17]. The Silhouette Coefficient is defined for each sample and is composed of two scores:
The mean distance between a sample and all other points in the same class. The mean distance between a sample and all other points in the next nearest cluster.

The value of the silhouette coefficient of a point varies between −1 and 1. A value near −1 indicates that the point is clustered badly. A value near 1 indicates that the point is well-clustered. To evaluate the quality of a clustering we can compute the average silhouette coefficient of all points.

$$s = \frac{b - a}{max(a, b)} \quad \ldots\ldots(1)$$

### 7.2 Event Clusters for Multiple News Article

Cluster 1 = 0.41, 0.46, 0.45
{agriculture, begin, examination}
Cluster 2 = 0.51, 0.52, 0.55, 0.57
{festival,diwali, pongal, christmas, marriage}
{general meeting} {foreign affairs} {incident} res
Cluster 3 = 0.66
{competition}
Cluster 4 = 0.70, 0.71, 0.72
{education, maintenance, war}
Cluster 5= 0.82, 0.85, 0.80
{complaint, order} {treatment}
{dance} res
Cluster 6 = 0.91, 0.9
{protection, election} respectively

Table 1 represents the silhouette coefficient for various sample points. A() and B() are the distance between the sample point and the various points within the same cluster and various points within the nearest cluster. This table represents evaluation for multiple news articles.

Table 1 : Sihouette Coefficient for various sample points

| SAMPLE POINT | A() | B() | SIL. COEFFICIENT |
|---|---|---|---|
| 0.45 | 0.025 | 0.093 | 0.731 |
| 0.57 | 0.036 | 0.045 | 0.2 |
| 0.66 | 0 | 0.62 | 1 |
| 0.72 | 0.02 | 0.06 | 0.66 |
| 0.82 | .0.02 | 0.1 | 0.8 |
| 0.91 | 0.05 | 0.09 | 0.8 |

In Fig.8 the value of the silhouette coefficient for multiple news article of a point varies between −1 and 1. A value near −1 indicates that the point is clustered badly. A value near 1 indicates that the point is well-clustered.

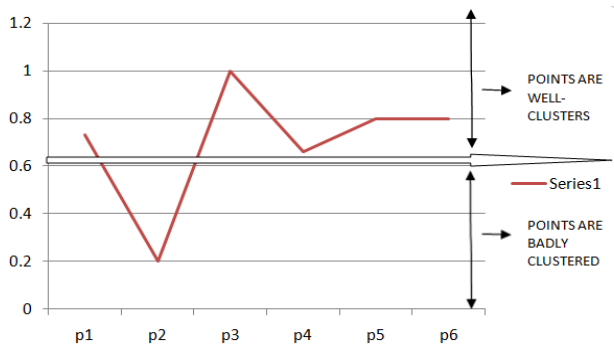

Fig.8.Context Specific Event Cluster Analysis

The silhouette coefficient is a measure for the clustering quality that is rather independent from the number of clusters. Experiences show that values between 0.7 and 1.0 indicate clustering results with excellent separation between clusters; *viz.* data points are very close to the center of their cluster and remote from the next nearest cluster. For the range from 0.5 to 0.7 one finds that data points are clearly assigned to cluster centers. Values from 0.25 to 0.5 indicate that cluster centers can be found, though there is considerable "noise", i.e. there are many data points that cannot be clearly assigned to clusters. Below a value of 0.25 it becomes practically impossible to find significant cluster centers and to definitely assign the majority of data points.

## 8. Conclusion

Event Specific sentences are extracted from the UNL Graph of sentences using the conditions. The conditions used for extracting event specific sentences can be modified that more conditions are added so that the efficiency of the proposed work can be further improved. The segmentation uses the scoring scheme in which the condition score and feature score can be further modified. Temporal information's are extracted perfectly by the proposed system.

The event weight score for computing the event context similarity between the documents is based on the similarity between the concept and its event specific UNL context. The event specific context has been identified by the word level semantics (semantic constraints), sentence level semantics (UNL relations exist between the concepts) and context level semantics (UNL attributes). However this approach uses graph based UNL event semantics, clustering specific event in a single cluster was difficult. Hence we have taken additional similarity features such as Time, Place and Persons similarity, instead of considering only UNL event semantics.

The similarity between two event contexts is based on the number of event arguments passing between them. Though we get specific event clusters in a single cluster, we find difficult in identifying sub events of the event in a single cluster. Hence, in order to improve our cluster efficiency, connective terms between two sentences are also important. Hence by combining UNL semantics, event specific argument's similarity between sentences produces good clusters. Identification sub-events are performed. Context specific Event Indexing and  Event Ranking performed based on new approach referred as modified scoring scheme.

## 9. Future Work

Scoring can be further improved for better results. In order to get more cohesive clusters we further extend our feature set into sentence level similarity and addition weight for connective terms between sentences. Generation of Domain specific event templates for the News Articles.